\documentclass[conference]{IEEEtran}
\IEEEoverridecommandlockouts
\usepackage{cite}
\usepackage{amsmath,amssymb,amsfonts}
\usepackage{algorithmic}
\usepackage{graphicx}
\usepackage{textcomp}
\usepackage{xcolor}
\usepackage{booktabs}
\usepackage{hyperref}
\usepackage{multirow}
\def\BibTeX{{\rm B\kern-.05em{\sc i\kern-.025em b}\kern-.08em
    T\kern-.1667em\lower.7ex\hbox{E}\kern-.125emX}}
\begin{document}

\title{LINR Bridge: Vector Graphic Animation via Neural Implicits and Video Diffusion Priors}

\author{\IEEEauthorblockN{Wenshuo Gao, Xicheng Lan, Luyao Zhang, Shuai Yang*}
\IEEEauthorblockA{Wangxuan Institute of Computer Technology, State Key Laboratory of Multimedia Information Processing, \\Peking University, Beijing, China}\vspace{-4mm}
\thanks{\footnotesize{$\ast$~Corresponding author. This work was supported in part by the National Natural Science Foundation of China under Grant 62471009, in part by CCF-Tencent Rhino-Bird Open Research Fund, in part by the Key Laboratory of Science, Technology and Standard in Press Industry (Key Laboratory of Intelligent Press Media Technology), and in part by The Fundamental Research Funds for the Central Universities, Peking University.}}
}

\newcommand{\ys}{\textcolor{magenta}}
\newcommand{\td}{\textcolor{red}}
\newcommand{\gws}{\textcolor{purple}}

\maketitle


\begin{abstract}
Vector graphics, known for their scalability and user-friendliness, provide a unique approach to visual content compared to traditional pixel-based images. Animation of these graphics, driven by the motion of their elements, offers enhanced comprehensibility and controllability but often requires substantial manual effort. To automate this process, we propose a novel method that integrates implicit neural representations with text-to-video diffusion models for vector graphic animation. Our approach employs layered implicit neural representations to reconstruct vector graphics, preserving their inherent properties such as infinite resolution and precise color and shape constraints, which effectively bridges the large domain gap between vector graphics and diffusion models. The neural representations are then optimized using video score distillation sampling, which leverages motion priors from pretrained text-to-video diffusion models. Finally, the vector graphics are warped to match the representations resulting in smooth animation. Experimental results validate the effectiveness of our method in generating vivid and natural vector graphic animations, demonstrating significant improvement over existing techniques that suffer from limitations in flexibility and animation quality.
\end{abstract}


\section{Introduction}
\label{sec1}

Vector graphics are a widely used form of flat visual content, distinct from traditional pixel-based raster images. They consist of defined elements such as circles, squares, lines, and Bézier curves, and are known for their simplicity, flatness, resolution-independent scalability, and user-friendliness. Consequently, vector graphics are extensively applied in icon design, animation, posters, and web design. Scalable Vector Graphics (SVG)~\cite{andersson2008scalable} is a popular XML-based format for representing them.

Unlike raster images, whose animations rely on pixel color changes, vector graphic animations are achieved by transforming graphic elements, offering greater clarity and control. However, producing smooth and coordinated motion often demands considerable manual effort. Typically, the process involves constructing a skeleton and applying transformations accordingly.

This highlights the need for automated vector graphic animation systems. Recently, diffusion-based video generation models such as ModelScope~\cite{wang2023modelscope}, VideoCrafter2~\cite{chen2024videocrafter2}, DynamiCrafter~\cite{xing2023dynamicrafter}, I2VGen-XL~\cite{zhang2023i2vgen}, SVD~\cite{blattmann2023stable}, and CogVideoX~\cite{yang2024cogvideox} have gained rapid traction. One approach is to render vector graphics into raster images and animate them using these pretrained models with text prompts. However, due to their training on natural imagery and lack of shape/color constraints, the results are often unstable and inconsistent with expectations~\cite{gal2024breathing, wu2024aniclipart}. Therefore, a dedicated system for vector graphic animation is essential.

Although some methods for vector graphic animation have been proposed~\cite{gal2024breathing, wu2024aniclipart}, they face notable limitations. LiveSketch~\cite{gal2024breathing} directly optimizes SVG point parameters to animate sketches, but it is limited to global transformations and point-level local shifts, offering little control over shapes and often causing structural distortion and unpredictable motion. AniClipart~\cite{wu2024aniclipart} extracts SVG skeletons and applies As-Rigid-As-Possible (ARAP) deformation for cartoon animation. However, it depends heavily on instance-specific keypoints and skeletal movement, enforcing overly rigid constraints that result in stiff animations.
Furthermore, both methods render parameterized SVGs into raster images before applying diffusion model optimization via video score distillation sampling (VSDS)~\cite{gal2024breathing}. This rasterization introduces a significant domain gap, complicating optimization and limiting the effectiveness of diffusion priors—ultimately preventing natural SVG animation.

This domain gap motivated our research. We observed that the implicit neural representation (INR)~\cite{poole2022dreamfusion, chen2021learning, thamizharasan2024nivel, mildenhall2021nerf}, which is introduced and widely adopted in image and 3D modeling, is highly effective in reconstructing and animating vector graphics. INR models an image using a neural network, typically taking coordinate position information as input and producing the corresponding color as output. INR has several advantageous characteristics. By using coordinates as input, INR constructs images with infinite resolution, which aligns well with the nature of SVG. Moreover, INR offers shape manipulability, significantly bridging the domain gap between parameterized SVG and rasterized images.
Our findings suggest that our color-wise layered implicit neural representation (LINR) can effectively reconstruct vector graphics, as they consist of various shapes, each assigned a specific color, enabling easier and more comprehensible animation.

\begin{figure*}[htbp]
\centering
\includegraphics[width=\textwidth]{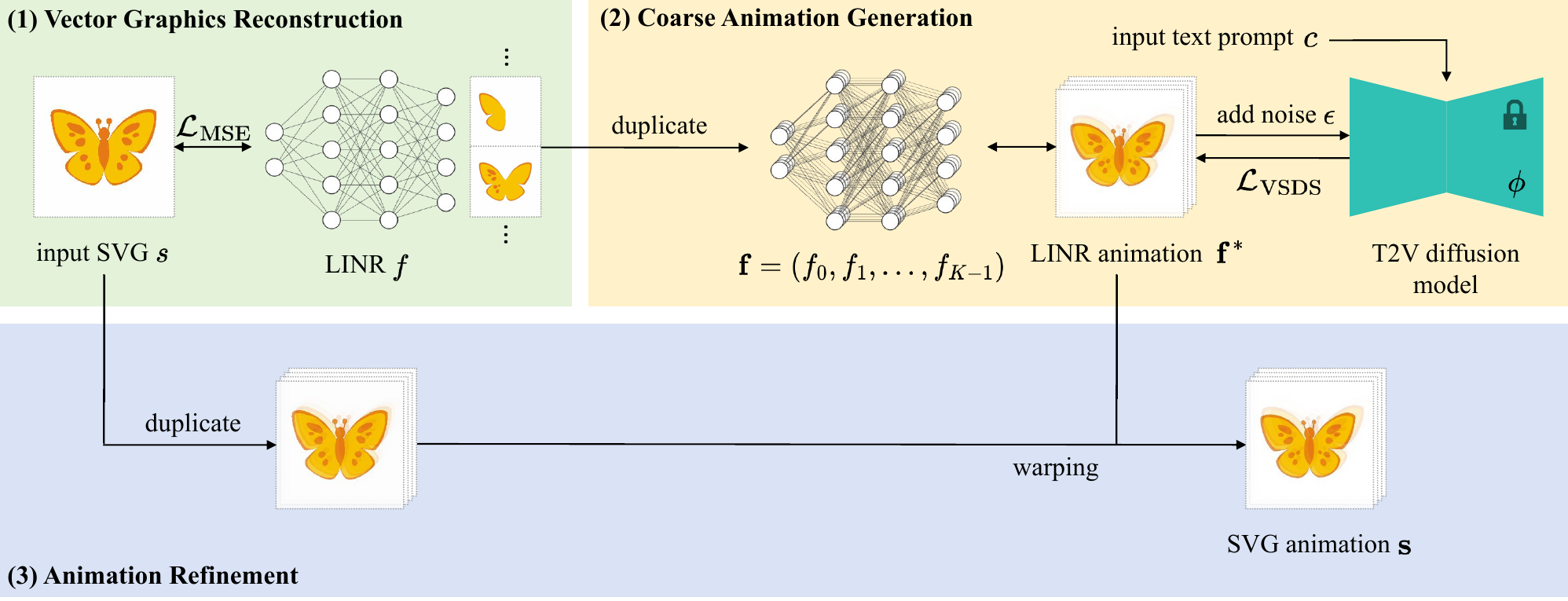}
    \caption{The pipeline of \textit{LINR Bridge} takes an SVG $s$ and a text prompt $c$ as inputs, and produces an SVG animation as output. The pipeline consists of three steps: (1) \textbf{Vector Graphics Reconstruction}: Optimize a LINR network $f$ to reconstruct the input SVG $s$. (2) \textbf{Coarse Animation Generation}: Replicate the network $f$ $K$ times to construct a $K$-frames static initial video $\mathbf{f}$. Input the frames rendered from $\mathbf{f}$ along with the text prompt $c$ into a text-to-video (T2V) diffusion model and optimize using VSDS to obtain a coarse LINR animation. (3) \textbf{Animation Refinement}: Warp $s$ based on optical flow or directly optimize $s$ to match the LINR animation, resulting in the final animated SVG $\mathbf{s}$.}
    \vspace{-5mm}
\label{fig1}
\end{figure*}


In this paper, we propose \textit{LINR Bridge}, a novel coarse-to-fine framework for vector graphic animation that utilizes LINR to bridge the domain gap between vector graphic representations and existing video diffusion models. Specifically, we first reconstruct the input vector graphics using LINR. This LINR is then replicated across the video frame dimension and optimized using VSDS to generate a coarse LINR animation. The input vector graphics are warped based on optical flow or directly optimized to match the LINR animation, resulting in the final vector graphic animation. LINR effectively constrains the color and shape of the images, while the VSDS loss introduces motion priors from the pretrained video diffusion models.
Experiments show that our method generates vivid and natural vector graphic animations.
Our contributions are summarized as follows:
\begin{itemize}
\item We propose an effective method using LINR to reconstruct vector graphics, resulting in high reconstruction quality, well-preserved shapes, and ease of optimization.
\item We combine LINR with VSDS for animating vector graphics. LINR serves as an intermediate representation to bridge the domain gap between parameterized SVG and raster images, empowering effective motion prior distillation from text-to-video (T2V) diffusion models.
\item By warping or optimizing SVG to match the intermediate LINR animation, we achieve complete and smooth SVG animation with appropriate constraints and flexibility. 
\end{itemize}
The rest of the paper is organized as follows. Sec.~\ref{sec2} introduces the proposed \textit{LINR Bridge}. Sec.~\ref{sec3} presents experimental results, and Sec.~\ref{sec4} provides  concluding remarks.

\section{Proposed Method}
\label{sec2}

Our method automates the generation of $K$-frame animations for a given SVG using a text prompt. This approach consists of three steps: (1) \textbf{Vector Graphics Reconstruction}: Optimize an LINR to reconstruct the input SVG; (2) \textbf{Coarse Animation Generation}: Replicate the LINR neural network $K$ times and optimize them with a text-to-video (T2V) diffusion model with VSDS loss, to generate a coarse LINR animation; (3) \textbf{Animation Refinement}: Warp or optimize the original SVG to match the LINR animation, resulting in a smooth SVG animation. The pipeline is illustrated in Fig.~\ref{fig1}.

\subsection{Vector Graphics Reconstruction}
Implicit neural representation (INR)~\cite{chen2021learning} involves using a neural network to represent image information. Typically, this involves building a neural network to map coordinate inputs $(X,Y)$ to color outputs $(R,G,B)$. Since the input coordinate is continuous, this representation produces an image with infinite resolution, which is an important characteristic of SVG. We follow the common practice to use SIREN~\cite{sitzmann2020implicit} as the neural network, which uses sine functions as activation functions to enhance the network's nonlinear representation ability and therefore image representation capability.
SVG consists of various elements each defined by its own color. To better fit this layered and flat nature, we are inspired by NIVeL~\cite{thamizharasan2024nivel} to replace the neural network outputs $(R, G, B)$ with $(m_1, m_2, \dots, m_L)$. Here, $L$ is the number of layers, $m_i \in [0,1]$ represents the coloring intensity for color $c_i$ at the $i$-th layer. Such a representation is called layered implicit neural representation
(LINR), which enhances the model's ability to handle layered graphics efficiently.


Specifically, based on the input SVG $s$, we extract its color information based on the stacking order of layers and merge adjacent regions with the same color. We denote the resulting color scheme as $(c_1, c_2, \dots, c_L)$, with $L$ representing the number of layers. We optimize an LINR network $f$ to map a coordinate input $\mathbf{p} = (X, Y)$ to an intensity vector $\mathbf{m} = (m_1, m_2, \dots, m_L)$, with $m_i \in [0,1]$. 
The process can be expressed as $\mathbf{m} = f(\mathbf{p})$.

We define $c_0$ as the background color and $c_i$ as the color for the $i$-th layer of the SVG. The RGB image $C(f)$ represented by a LINR $f$ is constructed as follows: For all coordinate points $\mathbf{p}$ in the canvas area, we first apply the background color $c_0$, then overlay each layer in order, with the $i$-th layer contributing color $c_i$ with intensity $m_i$. Here, an intensity of $0$ means that no color is applied and the layer is totally transparent at this point, while an intensity of $1$ means full color is applied and all colors in the lower layers are covered, and an intensity between $0$ and $1$ represents the opacity of the coloring in this layer, resulting in the RGB image $C(f)$.

The first stage optimizes LINR $f$ to reconstruct the input SVG $s$ with the following objectives:
\begin{itemize}
\item Layered reconstruction loss ($\mathcal{L}_{\text{MSE}}$): Measures the mean squared error between $C(f)$ and the raster image $R(s)$ rendered from $s$ among every position $\mathbf{p}$ and layer $i$, leading to $\mathcal{L}_{\text{MSE}} = \text{MSE}(R(s), C(f))=\mathbb{E}_{\mathbf{p},i}[(s(\mathbf{p})_i-f(\mathbf{p})_i)^2]$ where $s(\mathbf{p})_i \in \{0,1\}$ denotes if position $\mathbf{p}$ in layer $i$ of $s$ is colored, and $f(\mathbf{p})_i$ denotes the color intensity $m_i$ at position $\mathbf{p}$ and layer $i$.
We use differentiable rasterizer DiffVG~\cite{li2020differentiable} to obtain $R(s)$. 
\item Binarization loss ($\mathcal{L}_{\text{bi}}$): To maintain the sharp shapes of SVG, we would like the color intensity $m_i$ to approach the extreme values of $0$ or $1$. We penalize $B(m_i)=min\{(m_i-0)^{1.1},(1-m_i)^{1.1}\}$ at each position $\mathbf{p}$ and layer $i$, averaged over all positions and layers for $f$, 
leading to $\mathcal{L}_{\text{bi}} = \mathbb{E}_{\mathbf{p},i}[B(f(\mathbf{p})_i)]$. We choose this penalty because when $m_i$ is close to $0$ or $1$, the gradient approaches $0$, and in other regions, the gradient is approximately $1$ or $-1$. This behavior facilitates optimization and mitigates the risk of gradient explosion.
\item Regularization loss ($\mathcal{L}_{\text{reg}}$): Applied to network parameters to prevent gradient explosion with $\mathcal{L}_{\text{reg}} = L_2(f)$.
\end{itemize}
For reconstruction, we aim to optimize $f$ to obtain $f^* = \arg\min_f\mathcal{L}_{\text{rec}}$ with
\begin{equation}
\label{eq1}
    \mathcal{L}_{\text{rec}} = \mathcal{L}_{\text{MSE}} + \lambda_1 \mathcal{L}_{\text{bi}} + \lambda_2 \mathcal{L}_{\text{reg}},
\end{equation}
where $\lambda_1$ and $\lambda_2$ are weight parameters.

\textbf{Remark.} Our proposed LINRs align closely with key SVG characteristics:
\begin{enumerate}
    \item LINRs support infinite resolution, which aligns perfectly with the resolution independence of SVGs.
    \item LINRs have layered characteristics that match the natural hierarchical structure of SVGs.
    \item LINRs mimic the color filling techniques of SVGs for color consistency.
    \item LINRs have sharp shape boundaries, paralleling SVGs' ability to render crisp edges using mathematical curves.
\end{enumerate}
These four properties show how LINRs effectively exploit SVG traits, enabling superior performance in resolution-independent and animation-critical applications.

\subsection{Coarse Animation Generation}


Video score distillation sampling (VSDS) originates from score distillation sampling (SDS), which was first used to generate 3D models based on an image diffusion model~\cite{rombach2022high, poole2022dreamfusion, mildenhall2021nerf} and can be generalized to SVG~\cite{thamizharasan2024nivel, iluz2023word}. The core idea is to extract prior information from the image diffusion model by comparing the difference between the predicted noise and the added noise. Later, researchers extended image diffusion models to video diffusion models, proposing VSDS for video generation~\cite{gal2024breathing, wu2024aniclipart, liu2024dynamic}. However, the large domain gap between SVG and diffusion models makes it extremely hard to directly optimize. Thus, our key idea is to use LINR to bridge the gap.

We replicate the network $f^*$ $K$ times to get $\mathbf{f} = (f_0, f_1, \dots, f_{K-1})$ which constructs a static $K$-frame video $C(\mathbf{f})$. Next, we optimize $\mathbf{f}$ to produce a coarse LINR animation $\mathbf{f}^*$ with:
\begin{itemize}
\item VSDS loss ($\mathcal{L}_{\text{VSDS}}$): To ensure visual stability, we propose a new \textit{stability extension} method to extend $C(\mathbf{f})$ with a fixed frame of the original image at the beginning (and optionally at the end), resulting in a $(K + 1)$-frame video $V(\mathbf{f})=\text{concat}(R(s), C(\mathbf{f}))$. This fixed frame serves as an anchor to constrain the appearance of the remaining frames, which effectively ensures visual stability.
We then add noise $\epsilon$ to these frames at time step $t$ to get $V'(\mathbf{f})=\alpha_tV(\mathbf{f}) + \sigma_t\epsilon$ where $\alpha_t$ and $\sigma_t$ are pre-defined parameters depending on $t$ and use a T2V diffusion model $\epsilon_\phi$ with text prompt $c$ to predict noise $\epsilon_\phi(V'(\mathbf{f}), t, c)$. The VSDS loss is the prediction error: $\mathcal{L}_{\text{VSDS}} = \mathbb{E}_t[w(t)(\epsilon_\phi(V'(\mathbf{f}), t, c) - \epsilon)]$ where $w(t)$ is a constant depending on $\alpha_t$.
Note that we optimize LINR network parameters with the backpropagated VSDS gradients, which is inherently different from directly optimizing images. LINR parameters determine SVG position and shape, enabling natural motion learning, thanks to our LINR design that closely matches SVGs' characteristics.
\item Binarization loss ($\mathcal{L}_{\text{bi}}$): During the animation process, maintaining sharp edges of shapes is still important. We adopt binarization loss in Eq. (\ref{eq1}) by taking average among $K$ frames.
\item Regularization loss ($\mathcal{L}_{\text{reg}}$): Applied to network parameters to prevent gradient explosion. We adopt regularization loss in Eq. (\ref{eq1}) by taking average among $K$ frames.
\end{itemize}
For animation, our goal is to find $\mathbf{f}^* = \arg\min_\mathbf{f}\mathcal{L}_{\text{ani}}$ with
\begin{equation}
\label{eq2}
    \mathcal{L}_{\text{ani}} =  \mathcal{L}_{\text{VSDS}} + \lambda_3 \mathcal{L}_{\text{bi}} + \lambda_4 \mathcal{L}_{\text{reg}},
\end{equation}
where $\lambda_3$ and $\lambda_4$ are weight parameters, respectively.

\begin{figure*}[t]
\centering
\includegraphics[width=0.8\textwidth]{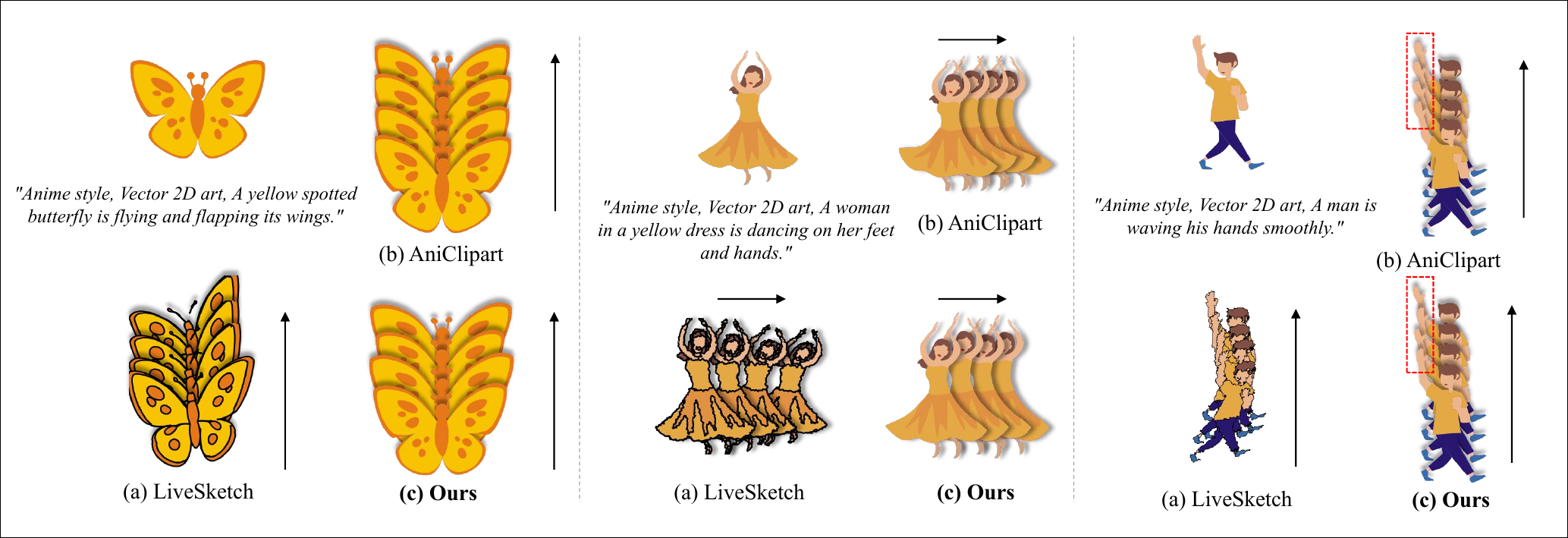}
    \caption{Comparisons of \textit{LINR Bridge} to other SVG animating methods. We show consecutive frames of SVG animation. (a) LiveSketch~\cite{gal2024breathing} optimizes global transforms and local point parameters, unsuitable for common vector graphics other than sketches. (b) AniClipart~\cite{wu2024aniclipart} optimizes an ARAP deformation on vector graphics and offers excessively strict restrictions on shapes. (c) Our method can generate smooth, highly flexible, and well-structured animation results.} 
    \vspace{-5mm}
\label{fig2}
\end{figure*}

\subsection{Animation Refinement}
After obtaining the LINR animation $\mathbf{f}^*$, we warp the original SVG parameters to match $C(\mathbf{f}^*)$. We first calculate the optical flows from $R(s)$ to each frame in $C(\mathbf{f}^*)$ using Farneback method~\cite{farneback2003two}. Then we apply the optical flows to every point parameters in $s$ to produce a smooth SVG-based animation $\mathbf{s}$.

For scenarios that require more flexibility and freedom, such as multi-layer ones, we provide an alternative approach: directly optimizing the point parameters of $s$ using MSE loss with each frame of LINR animation $C(\mathbf{f}^*)$ to produce $\mathbf{s}$.
%
First, we duplicate $s$ to obtain $\mathbf{s} = (s_0, s_1, \dots, s_{K-1})$ and then optimize $\mathbf{s}$ with the objectives:
\begin{itemize}
\item Matching loss ($\mathcal{L}_{\text{MSE}}$): Mean squared error loss between the frames $R(\mathbf{s})$ differentiably rendered~\cite{li2020differentiable} from $\mathbf{s}$ and the frames $C(\mathbf{h^*})$ rendered from the corresponding $\mathbf{h^*}$: $\mathcal{L}_{\text{MSE}} = \text{MSE}(R(\mathbf{s}), C(\mathbf{h}^*))$.
\item Structural loss ($\mathcal{L}_{\text{str}}$): 
We follow~\cite{iluz2023word,liu2024dynamic} to penalize the difference in triangular mesh-based structure between the optimized SVG and the original SVG, as well as between adjacent SVG frames to increase structural consistency and uniformity of the SVG in the process of optimization.
We denote this loss as $\mathcal{L}_{\text{str}} = S(\mathbf{s})$.
\end{itemize}
Thus, we seek $\mathbf{s}^*=\arg\min_\mathbf{s} \mathcal{L}_{\text{Fine}}$ with
These losses are combined with a weight parameter $\lambda_5$.
Our goal is to optimize $\mathbf{s}$ to minimize the losses:
\begin{equation}
\label{eq3}
    \mathcal{L}_{\text{Fine}} = \mathcal{L}_{\text{MSE}} + \lambda_5 \mathcal{L}_{\text{str}},
\end{equation}
where $\lambda_5$ is a weight parameter.

\section{Experiments}
\label{sec3}

\subsection{Experiment Setup}
The LINR network is constructed by SIREN network~\cite{sitzmann2020implicit} which has $4$ layers with a hidden dimension of $64$ and is activated by a clip-sigmoid function where 
$y = \text{clamp}_{[0,1]}\left((\text{sigmoid}(x) - 0.5) \times (1 + 10^{-7}) + 0.5\right)$ to prevent large parameters and gradient explosion since a sigmoid function is not able to approach the extreme values of $0$ or $1$. The differentiably rendered image size is $256 \times 256$ pixels. 

During \textbf{Vector Graphics Reconstruction}, $\lambda_1$ increases linearly from $0$ to $2$ over the optimization process—prioritizing overall consistency early and sharp edges later. $\lambda_2$ is fixed at $1 \times 10^{-6}$. We use the Adam optimizer~\cite{kingma2014adam} with a learning rate of $1 \times 10^{-3}$ for 10,000 iterations.

For \textbf{Coarse Animation Generation}, a 12-frame video is rendered using the ModelScope text-to-video model~\cite{wang2023modelscope}, with timestep $t$ randomly sampled between 200 and 400 each iteration. $\lambda_3$ and $\lambda_4$ are set to $2 \times 10^2$ and $1 \times 10^{-6}$, respectively. Adam is used with a learning rate of $3 \times 10^{-5}$ for 8,000 iterations, saving results every 100 iterations to obtain 80 intermediate LINR animations.

In \textbf{Animation Refinement}, $\lambda_5$ is set to $2 \times 10^{-7}$. Adam is used with a learning rate of $5 \times 10^{-1}$ for 400 iterations, incorporating intermediate results from the next coarse LINR animation every 5 iterations.

\subsection{Metrics}

We collected approximately 80 SVG vector graphics, covering about 18 human, 24 animals, 6 plants, 22 objects and 10 icons, animated them, and measured the animation results in terms of \textbf{Appearance Consistency} and \textbf{Motion-Prompt Alignment}. The results are shown in Table~\ref{tab1}. For \textbf{Appearance Consistency}, we measured the CLIP~\cite{radford2021learning} cosine similarity between the generated frames and the original image. For \textbf{Motion-Prompt Alignment}, we measured the X-CLIP~\cite{ni2022expanding} similarity score between the generated video and the input text prompt.

\begin{figure*}[ht]
\centering
\includegraphics[width=0.63\textwidth]{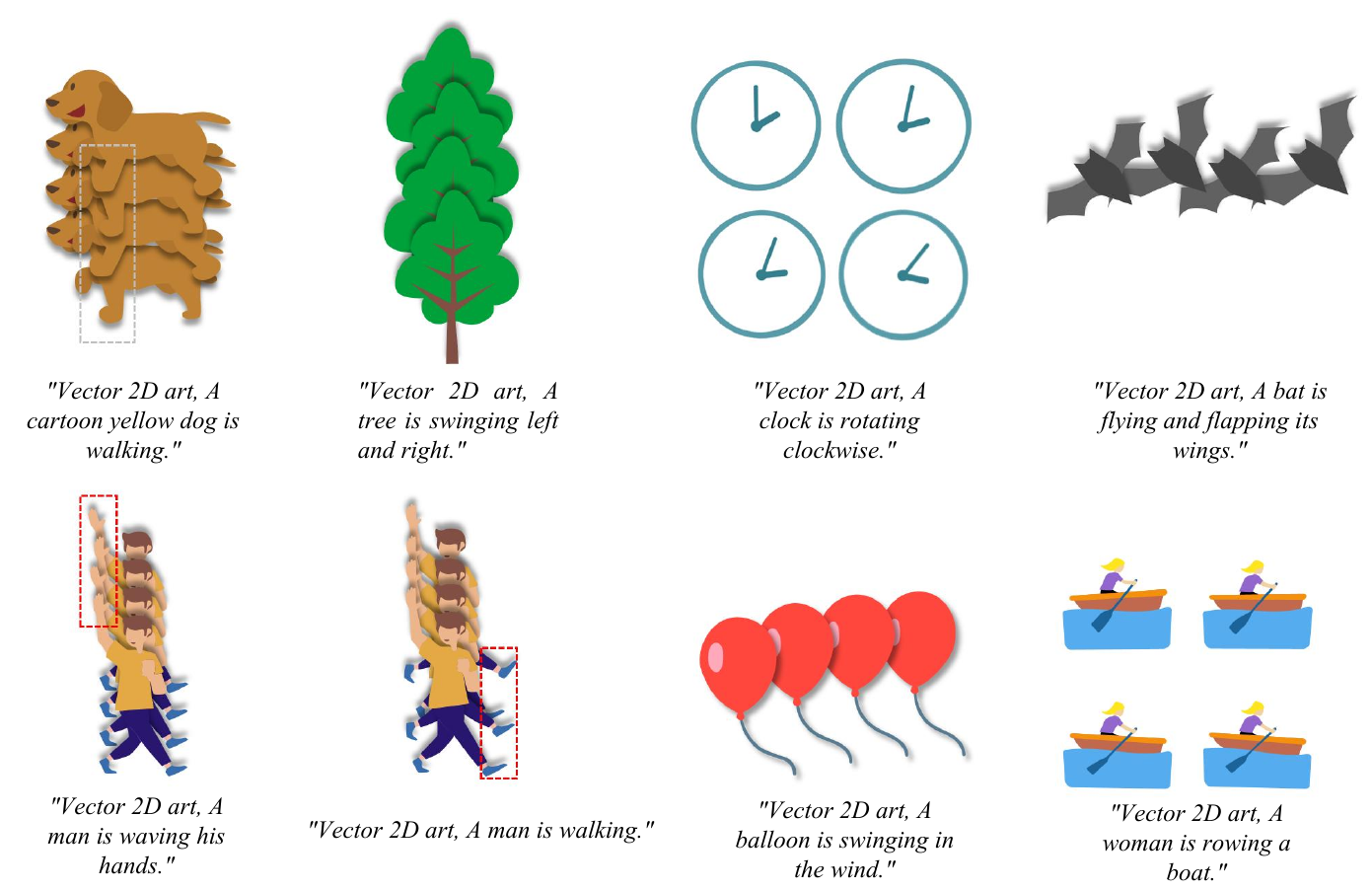}
    \caption{Validating diversity and different prompts. Consecutive frames are selected from animation generated by different prompts.}
\vspace{-5mm}
\label{fig4}
\end{figure*}

\begin{figure}[h]
\centering
\includegraphics[width=\columnwidth]{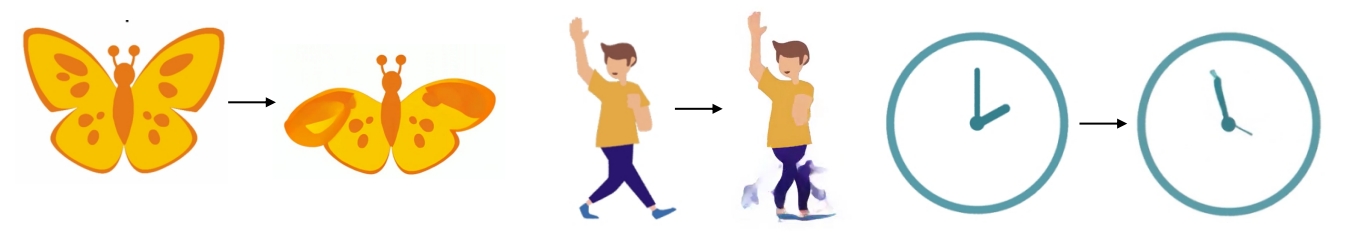}
\caption{Failure cases of direct using of video diffusion models for SVG animation. The leftmost showcase is from SVD~\cite{blattmann2023stable} and the others are from CogVideoX~\cite{yang2024cogvideox}.}
\vspace{-5mm}
\label{fig:fc}
\end{figure}

\subsection{Comparison with State-of-the-Art Methods}

Given the scarcity of automated vector graphic animation methods, we compare our approach with the most related two methods. Example results are shown in Fig.~\ref{fig2}.

\begin{itemize}
\item Comparison with Direct Use of Video Diffusion Models: Methods like ModelScope~\cite{wang2023modelscope}, VideoCrafter2~\cite{chen2024videocrafter2}, DynamiCrafter~\cite{xing2023dynamicrafter}, I2VGen-XL~\cite{zhang2023i2vgen}, SVD~\cite{blattmann2023stable}, and CogVideoX~\cite{yang2024cogvideox} rely on video diffusion models trained on natural image domains, lacking constraints on color and shape. Consequently, results often show significant deviations in shape and color, with issues such as distortions and artifacts, shown in Fig.~\ref{fig:fc}.

\item Comparison with LiveSketch~\cite{gal2024breathing}: LiveSketch is designed for animating sketches. It focuses on global and local optimization by inputting the SVG's parameter points into an MLP to obtain optimized values, which are then animated using VSDS which leads to shape alterations, unsmooth lines, and excessive freedom. The results are shown in Fig.~\ref{fig2}(a).

\item Comparison with AniClipart~\cite{wu2024aniclipart}: AniClipart extracts key points and skeletons from SVGs, performs As-Rigid-As-Possible (ARAP) deformations based on these key points, and animates using VSDS. SVG motion is constrained by skeleton movements, leading to limited flexibility and excessive joint motion. The results are shown in Fig.~\ref{fig2}(b).
\end{itemize}
In contrast, our method animates within the LINR domain, ensuring better consistency in color and shape, while providing greater smoothness, freedom of motion, and separation over layers. The results are shown in Fig.~\ref{fig2}(c). To further validate the ability of our method in various types of SVGs and different prompts, more results are shown in Fig.~\ref{fig4}.

\begin{table}[ht]
\centering
\caption{Comparisons to Other Methods}
\label{tab1}
\resizebox{\linewidth}{!}{
\begin{tabular}{c|c|c}
\toprule
\multirow{2}*{Method} & \textbf{Appearance Consistency} & \textbf{Motion-Prompt Alignment} \\
& $\uparrow$ & $\uparrow$ \\
\midrule
SVD & 0.8964 & 18.9659 \\
CogVideoX & 0.9181 & 20.5289 \\
LiveSketch & 0.8618 & 19.2401 \\
AniClipart & 0.9476 & 19.2327 \\
\textbf{Ours} & \textbf{0.9649} & \textbf{21.5139} \\
\bottomrule
\end{tabular}}
\end{table}

\begin{figure}[htbp]
\centering
\includegraphics[width=0.6\columnwidth]{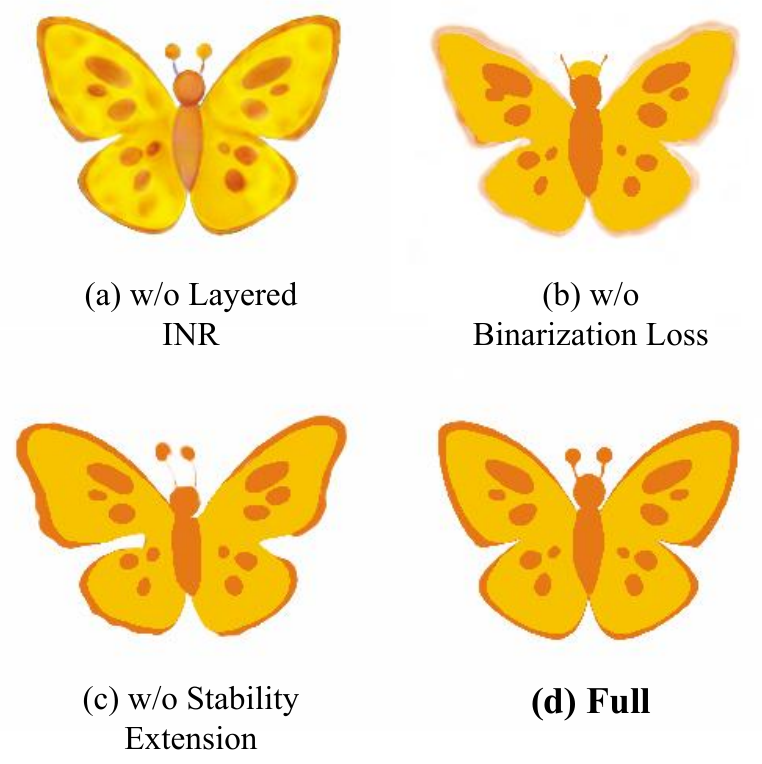}\vspace{-1mm}
    \caption{Ablation study. We show one frame from coarse LINR animations. (a) Without the layered structure of INR, the colors will become uncontrollable, leading to significant color deviations. (b) Without binarization loss, the rendered image will lack clear boundaries, resulting in blurred transition areas. (c) Without the stability extension, the content of the image will not be constrained, leading to a gradual deviation in appearance from the original. (d) Our full model generates the most visually appealing results.}
    \vspace{-5mm}
\label{fig3}
\end{figure}

\subsection{Ablation Study}

\begin{itemize}
\item Layered INR: The layered structure preserves each layer's shape simplicity and color consistency. Significant color deviations and shape changes may occur during VSDS if INR outputs $(R, G, B)$ values. See Fig.~\ref{fig3}(a).

\item Binarization Constraint: The binarization loss ensures clear edges in each layer of the LINR-generated image. LINR may produce blurry images and excessive shape variations during animation without this loss. See Fig.~\ref{fig3}(b).

\item Structural Constraint: The structural loss maintains the structural integrity and consistency of the SVG during optimization. Issues like unsmooth lines and excessive structural differences between adjacent frames may occur without this loss.

\item Stability Extension: The frame extended maintains the appearance consistency between the optimized LINR and the original SVG. Animation might gradually deviate from the shape of the original during continuous optimization without it. See Fig.~\ref{fig3}(c).
\end{itemize}

\section{Conclusion}
\label{sec4}
In this paper, we propose an automated method for vector graphics animation using LINR and VSDS. Our approach first reconstructs the input SVG using LINR and then generates animations by leveraging prior knowledge from T2V diffusion models through VSDS. Our method maintains consistency in shape and color while offering a high degree of freedom in motion, resulting in stable and smooth animations.

\bibliographystyle{IEEEtran}
\bibliography{sample-now}

\end{document}